\pdfoutput=1

\documentclass[11pt]{article}

\usepackage[final]{acl}

\usepackage{times}
\usepackage{latexsym}

\usepackage[T1]{fontenc}

\usepackage[utf8]{inputenc}

\usepackage{microtype}

\usepackage{inconsolata}

\usepackage{algorithm}
\usepackage{algpseudocode}
\usepackage{booktabs}

\usepackage{pgfplots}
\pgfplotsset{width=10cm,compat=1.9}

\usepackage{fontawesome5}

\usepackage{listings}
\title{LAR-ECHR: A New Legal Argument Reasoning Task and Dataset for Cases of the European Court of Human Rights}

\author{ Odysseas S. Chlapanis$^{1, 3}$, \hspace{0.5mm} Dimitrios Galanis$^{2, 3}$ \and  Ion Androutsopoulos$^{1, 3}$ \\
$^{1}$Department of Informatics, Athens University of Economics and Business, Greece\\
$^{2}$Institute for Language and Speech Processing,  Athena Research Center, Greece\\$^{3}$Archimedes Unit, Athena Research Center, Greece\\ 
\texttt{odyhlapanis@aueb.gr, galanisd@athenarc.gr, ion@aueb.gr}
}

\begin{document}
\maketitle

\begin{abstract}
We present Legal Argument Reasoning (LAR), a novel task designed to evaluate the legal reasoning capabilities of Large Language Models (LLMs).
The task requires selecting the correct next statement (from multiple choice options) in a chain of legal arguments from court proceedings, given the facts of the case. We constructed a dataset (LAR-ECHR) for this task using cases from the European Court of Human Rights (ECHR). We evaluated seven general-purpose LLMs on LAR-ECHR and found that (a) the ranking of the models is aligned with that of LegalBench, an established US-based legal reasoning benchmark, even though LAR-ECHR is based on EU law, (b) LAR-ECHR distinguishes top models more clearly, compared to LegalBench, (c) even the best model (\texttt{GPT-4o}) obtains 75.8\% accuracy on \mbox{LAR-ECHR}, indicating significant potential for further model improvement. The process followed to construct LAR-ECHR can be replicated with cases from other legal systems.
\end{abstract}

\section{Introduction}

The rise of Large Language Models (LLMs) has impacted several sectors, including the legal one. In the United States, LLMs are being integrated into legal research and writing tools designed for both professionals and laypeople.\footnote{\url{https://www.americanbar.org/groups/law_practice/resources/tech-report/2023/2023-artificial-intelligence-ai-techreport/}}
These advances are largely due to the effort of legal experts who contributed significantly in dataset development and manual evaluation \citep{guha2023legalbench, magesh2024hallucinationfreeassessingreliabilityleading}. Their involvement, however, is extremely costly, hence methods to construct and evaluate legal benchmarks semi-automatically are required. 

LegalBench \citep{guha2023legalbench} is an example of a widely used legal reasoning benchmark. It consists of data for 162 tasks, hand-crafted by legal experts, that evaluate six types of legal reasoning of the US legal system, making it the most reliable dataset of this kind. MMLU-Law, a subset of MMLU \citep{hendryckstest2021} that contains three US legal tasks only, is also commonly used. Similarly, LawBench \citep{fei2023lawbench} and IL-TUR \citep{joshi-etal-2024-il} were created for the Chinese and Indian regions; they comprise 20 and 8 tasks, respectively. For other legal systems, at least two other large multi-task legal benchmarks have been made available \citep{chalkidis-etal-2022-lexglue, niklaus-etal-2023-lextreme}. However, they include mostly text classification tasks that do not require understanding or generating chains of legal arguments (e.g., court arguments explaining the decisions of judges) and can be solved reasonably well using smaller BERT-based models \citep{chalkidis-etal-2020-legal} or even linear classifiers \citep{Aletras2016PredictingJD}. Hence, it is questionable if they test legal reasoning abilities. 

\begin{table}
  \centering
  \small
  \renewcommand{\arraystretch}{1.2} 
  \begin{tabular}{|l|}
    \hline
    \textbf{Facts:}\\
    Applicant married B, had daughter C. \\
    B abused applicant due to psychiatric disorders.\\
    B arrested, released; applicant moved to shelter. \\
    Marriage dissolved; B continued harassment.\\
    
    \textbf{Preceding arguments:}\\
    Government claimed it has established legal protections\\
    for domestic violence victims. The court noted B's\\
    repeated violence. All incidents concerned the same\\
    perpetrator and occurred in a continual manner, so the\\
    Court will examine them as a continuous situation.\\
    
    \textbf{Continuation Options:}\\
    (A) Psychiatric reports indicating mental disorders,\\
    tendency towards violence ... \hfill \faCheck \\
    (B) Applicant's confinement in mental hospital ...\hfill \faTimes {\ } \\
    (C) Police collected information about applicant ...  \hfill \faTimes {\ } \\
    (D) Complaint about psychiatric examination, ... \hfill \faTimes {\ } \\
    \hline
  \end{tabular}
  \caption{A cropped instance from LAR-ECHR.}
  \label{tab:example}
\end{table}

resent
Motivated by the observations above, we introduce a novel task (\textsection \ref{sec:task}), Legal Argument Reasoning (LAR), designed to evaluate the legal reasoning skills of LLMs. The task requires selecting the correct next statement (from multiple choice options) in a chain of legal arguments from court proceedings, given the facts of the case. We have prepared a LAR dataset for EU law, LAR-ECHR (\textsection \ref{sec:dataset}), using court arguments from the European Court of Human Rights (ECHR). It contains 403 instances; Table \ref{tab:example} shows a cropped example. Notably, the same process could be applied to construct LAR datasets for other legal systems as well.

We evaluate two closed-weight and 5 open-weight LLMs on LAR-ECHR (\textsection \ref{sec:experiments}), comparing their performance and rankings against two legal benchmarks: LegalBench and MMLU-Law. For completeness, we also report scores on two general reasoning benchmarks: the entire MMLU (MMLU-Full) and HellaSwag \citep{zellers-etal-2019-hellaswag}.
We find that: (a) the ranking of the models on LAR-ECHR is aligned with that of LegalBench, even though the two benchmarks are from different legal systems (US, EU); (b) LAR-ECHR provides clearer distinctions among top models, compared to LegalBench;
(c) even the best model (\texttt{GPT-4o}) obtains only 75.8\% accuracy on LAR-ECHR, similar to the top accuracy on LegalBench (73.3\%), indicating significant potential for further model improvement. 

Our contributions are three-fold: (i) we introduce a novel task, Legal Argument Reasoning (LAR), to evaluate the legal reasoning abilities of state-of-the-art LLMs; (ii) we create and publicly release LAR-ECHR, a challenging EU-specific LAR dataset; (iii) we present a method to create LAR datasets for any other legal system using court proceedings with annotated arguments.

\section{The LAR task} \label{sec:task}

We introduce Legal Argument Reasoning (LAR), a novel task to evaluate the legal reasoning abilities of LLMs. The LLM is fed with the \emph{facts} of the case (a list of sentences summarizing the events considered, see Table~\ref{tab:example}), a sequence of \emph{preceding arguments} (statements) from the court proceedings, and \emph{continuation options} (the correct next statement and distractors). The LLM has to select the correct next statement. (In court proceedings, `arguments' are numbered statements documenting the legal reasoning of the court.) 

LAR requires various types of legal and commonsense reasoning that extend beyond simple pattern recognition or memorization. As demonstrated by HellaSwag \citep{zellers-etal-2019-hellaswag}, even predicting the next sentence in a generic corpus can be challenging and requires reasoning skills. In the legal domain, the complex terminology and inferences make the task of predicting a legal professional's next argument significantly harder (\textsection \ref{sec:experiments}). The arguments embody the court's legal reasoning for its decisions. As \citet{medvedeva-mcbride-2023-legal} state: ``[J]udges usually offer explanations which serve to justify their decisions with reference to the facts found to be established and the relevant law. [These explanations] involve the exercise of legal reasoning''. Merely understanding the legal terminology is insufficient, as the the facts and the relevant law must also be considered.

\section{The LAR-ECHR dataset} \label{sec:dataset}
 
The LAR-ECHR dataset contains arguments from the proceedings of the European Court of Human Rights (ECHR). 
An ECHR court decision typically begins with the facts of the case, followed by the `Law' section, where the arguments of the parties and the court are presented, followed by the court's conclusion (e.g., verdict, fines). To create the dataset, we used statements from `Law'.

To ensure that LAR-ECHR is challenging and effectively evaluates legal reasoning, we select appropriate arguments based on criteria derived from the annotations of the Legal Argument Mining ECHR (LAM:ECHR) dataset \citep{habernal_mining_2024} (\textsection \ref{sec:data}). The criteria we use are described in \textsection \ref{sec:candidates} below.
Instead of generating the distractors using a language model, as in HellaSwag \citep{zellers-etal-2019-hellaswag} and LegalLens \citep{bernsohn-etal-2024-legallens}, we draw them from similar ECHR cases using an algorithm we developed (\textsection \ref{sec:distractors}), to avoid introducing language model biases and hallucinations.

LAR-ECHR is based on 191 ECHR court cases. From the 191 cases, we derive 403 samples (like the example of Table~\ref{tab:example}), which we split randomly into three sets: 5 samples for few-shot prompts, 98 samples for development, 300 samples for testing. In our experiments, we use only the test set, but we release the full dataset for future research.\footnote{\url{https://huggingface.co/datasets/AUEB-NLP/lar-echr}}
Below we describe in detail how the dataset was created.

\subsection{The previous LAM:ECHR dataset} \label{sec:data}

In the aforementioned LAM:ECHR dataset, the arguments of 373 ECHR court decisions were annotated for legal argument mining purposes.
The cases pertain to alleged violations of Article 8 (`Respect for private and family life') and, to a lesser extent, Article 7 (`No punishment without law') of the European Convention on Human Rights. The arguments were annotated for the actors stating them (`ECHR', `Applicant', `State', `Third Parties', `Commission/Chamber') and the type of argument (16 types). The argument types are: `Procedural', `Interpretation' (five variations), `Principle of proportionality' (four variations), `Institutional' (three variations), `Precedents', `Decision', `Application to the concrete case'. The latter type is the most common (57\%), and we use only arguments of this type in the new LAR-ECHR dataset (\textsection \ref{sec:candidates}).

There are 9,950 arguments (65\%) labeled with the `ECHR' actor in LAM:ECHR, 2,471 (16\%) arguments labeled with `Applicant', 2,399 (16\%) labeled with `State'. 
Only the remaining 385 (3\%) arguments are associated with the last two actors; for simplicity, we discard these 385 statements. 

The facts of each case are not included in LAM:ECHR, but they are included in the \mbox{ECtHR~B} dataset \citep{chalkidis-etal-2021-paragraph}, which does not provide arguments. We unified the two datasets  using regular expressions. A further complication is that LAM:ECHR was published after ECtHR B. The 373 cases of LAM:ECHR include 94 cases that are not covered by ECtHR B. Consequently, we used only the 279 cases covered by both datasets. Recently, a new dataset, ECtHR-PCR \citep{t-y-s-s-etal-2024-ecthr}, which contains both facts and arguments of ECHR cases (even the most recent ones), was released. Using this dataset, the missing cases will be included in an update of LAR-ECHR in the future.

From the remaining 279 cases, we selected the most appropriate \emph{target arguments} (correct next statements) according to criteria described in \textsection \ref{sec:candidates} below. Some cases included multiple arguments that satisfied the criteria, while others none (88 cases). Consequently, we selected target arguments from the remaining 191 cases. The distractor arguments (incorrect next statements) were also selected from the 191 cases (\textsection \ref{sec:distractors}). This process led to 403 instances, like the one of Table~\ref{tab:example}.

\subsection{Selection of target arguments}
\label{sec:candidates}
Here we describe the process used to select the target arguments (correct next statements) of the new dataset from the 191 cases of \textsection \ref{sec:data}. 

As already noted, the `Law' section of each case contains the arguments of the parties and the court. Actually, a case usually examines multiple \emph{issues} and the `Law' section contains the arguments of the parties, followed by the arguments of the court, separately per issue. For each issue, the arguments of the parties (in the court proceedings) are actually also written by the judges, in a way that supports the reasoning of the judges. Therefore, for each issue, the arguments (statements) of both the parties and the court actually form a reasoning chain. From that chain, we wish to focus on the arguments of the judges, especially those annotated as `Application to the concrete case' in LAM:ECHR, which are the most demanding in terms of reasoning, as they consider and combine the arguments of the parties, the law, and the facts of the particular case. Those arguments are ``concerned with determining the relation between the concrete case and the abstract legal norm by the subsumption of the facts of a case under a legal norm'' \citep{habernal_mining_2024}.
In other words, they are parts of the reasoning that the judges follow to connect the law to the facts by `subsumption', i.e., checking if the facts meet the conditions specified by the law.

Furthermore, in our experience, among the arguments of the judges, the first one (per issue) is the most difficult to predict; we leave an experimental validation of this claim for future work (\textsection \ref{sec:future}). Therefore, we select as target arguments those that satisfy the following criteria: (i) the argument must be annotated as `ECHR' (argument of the judges), (ii) the argument must be annotated as `Application to the concrete case', and (iii) it must be the first one (per issue) after the arguments of the parties. 

Due to the limited context length of LLMs, in LAR-ECHR the facts of each case are summarized (using \texttt{GPT-4o}) and only the last three of the arguments preceding the target one are retained. 

\subsection{Selection of distractors}
\label{sec:distractors}

Distractors are incorrect next statements, as opposed to the target argument, which is the correct one.
Some studies use synthetic distractors generated by LLMs, e.g., HellaSwag \citep{zellers-etal-2019-hellaswag} and Legalens \citep{bernsohn-etal-2024-legallens}.
We opt to use arguments from the same dataset as distractors, following the approach in EntailmentBank \citep{dalvi-etal-2021-explaining}. 
This approach avoids the introduction of biases and hallucinations of LLM generators, as reported in the work of HellaSwag. 

The most suitable distractors are algorithmically selected. 
The algorithm adheres to the following desiderata. 
(a) The distractors must be \textit{similar} to the target argument, i.e., they must have roughly the same style, length, and vocabulary. As seen in the example in Table \ref{tab:example}, the target (correct) argument refers to an event about `psychiatric reports', `mental disorder', and `tendency towards violence'; each one of the distractors mentions relevant terms (`mental hospital', `police', `psychiatric examination').
However, (b) the distractors should not be near-duplicates or paraphrases of the target argument or another distractor. With these desiderata, we developed the following algorithm.


\begin{table}
  \centering
  \small
  \renewcommand{\arraystretch}{1.1} 
  \begin{tabular}{lc}
    \hline
    \textbf{Text} & \textbf{Score} \\
    \hline
    \hline
    \textbf{Target argument}\\
    \hline
    \hline
    The Court notes that this complaint & \\ 
    is not manifestly ill-founded within &  1.00\\ 
    the meaning of Article 35 \textsection 3 ...&\\ 
    \hline
    \hline
    \textbf{Candidate distractors}\\
    \hline
    \hline
    The Court notes that \textit{the application} & \\ 
    is not manifestly ill-founded within &  0.95\\ 
    the meaning of Article 35 \textsection 3 ...& \\ 
    \hline
    The Court, having examined those & \\
    complaints under Articles 5 \textsection 1 & 0.85\\
    and 6 \textsection 1 of the Convention ...& \\
    \hline
    The Court considers that this part& \\
    of the application raises questions& \textbf{0.79}\\
    of law which are important ...& \\
    \hline
    The Court notes that the Government & \\
    put forward reasons for this complaint & 0.73\\
    to be declared inadmissible ...& \\ 
    \hline
  \end{tabular}
  \caption{Exploring the effect of the cosine similarity threshold $\tau$ on the candidate distractors.}
  \label{tab:threshold}
\end{table}

\smallskip
\noindent 
\textbf{Distractor selection algorithm:} For each target argument, the candidate distractors are the target arguments of the other cases (of all the issues of the other 190 cases, \textsection \ref{sec:data}). For each candidate distractor, its embedding is computed using an LLM.\footnote{We use \texttt{openai-embed-small}  
(\url{https://openai.com/index/introducing-text-and-code-embeddings/})
\citep{neelakantan2022textcodeembeddingscontrastive}.}
The candidate distractors are then ranked based on their cosine similarity to the embedding of the target argument (desideratum (a)), from highest to lowest.
While the top-3 ranked candidates could present the greatest challenge, they may also be paraphrases of the target argument or another distractor (desideratum (b)). Hence, before selecting the top-3 ranked candidate distractors, we discard candidate distractors whose similarity to the target argument or a more highly ranked candidate distractor exceeds a threshold $\tau$. 

\smallskip
\noindent
\textbf{Cosine similarity threshold:} To select the $\tau$ threshold, we conducted the following experiment: for each one of a few target arguments of the development subset (\textsection \ref{sec:setup}), we ranked the candidate distractors as above, and manually inspected the texts of the target and the distractors and their similarity scores (see Table~\ref{tab:threshold} for an example).
We observed that for similarity scores above 0.9, the two texts were almost identical. For scores between 0.9 and 0.85, they shared several words. For lower similarity scores, no such issues were visible, so we set $\tau = 0.8$. 

\begin{table*}[t]
  \centering
  \begin{tabular}{l|ccc|cc}
    Models & \, LAR-ECHR  & Legal & MMLU \; &  \; MMLU&  \; Hella\\
    & {(Ours)} & Bench* & Law* & \; Full*&  \; Swag\\
    \hline
    GPT-4o (L) \, & \, {\textbf{75.8} $\pm$ 1.8 [\textbf{1}]} & {\textbf{73.3}}   [\textbf{1}]& {\textbf{85.2}}  [\textbf{1}] & \textbf{74.8}   [\textbf{1}]& \textbf{89.1}  [\textbf{1}]\\
    GPT-4o-mini (S) & \, {61.6 $\pm$ 2.2}  [4]& {65.3}   [4]& {79.6}  [2] & 66.8  [4]& {83.4} [3]\\
    \hline
    Mistral-8x22B (L) \,& \, {69.8 $\pm$ 1.3}   [2]& {70.8}   [2]& {79.1}  [3] & {70.1}  [3]& {79.6}  [4]\\
    Mistral-8x7B (M) & \, {57.2 $\pm$ 1.6}   [5]& {63.0}   [5]& {74.3}  [4] & {64.9}  [5]& {70.5}  [5]\\
    Mistral-7B (S) & \, {49.6 $\pm$ 1.9}  [7] & {33.1}   [7]& {63.2}  [6] & {58.4}  [6]& {60.7}  [7]\\
    \hline
    Llama-3.1-70B (L) \,& \, {67.2 $\pm$ 2.6}   [3]& {68.7}   [3]& {67.4}  [5] & {70.9}  [2]& {86.2} [2]\\
    Llama-3.1-8B (S) & \, {54.1 $\pm$ 1.6}   [6]& {34.2}  [6] & {57.3}  [7] & {50.0}  [7]& {68.0}  [6]\\
    \hline
  \end{tabular}
  \caption{\label{tab:main}
    Comparison of LLMs from three families on LAR-ECHR, LegalBench, MMLU-Law, MMLU-Full. L, M, S denote the largest, medium, smallest models per family, respectively. Accuracy in \%. Rankings in square brackets. LAR-ECHR results averaged over three random seeds. HellaSwag results on a subset of 1000 samples and one random seed. *Results from \citet{liang2023holisticevaluationlanguagemodels}.
  }
\end{table*}
\section{Experiments}
\label{sec:experiments}
\subsection{Experimental setup}
\label{sec:setup}

We evaluate the reasoning skills of seven general-purpose LLMs using the respective web APIs and three random seeds.
We employed closed-weight OpenAI models (GPT family), namely \texttt{gpt-4o} (L), \texttt{gpt-4o-mini} (S) \citep{openai2024gpt4technicalreport}\footnote{\url{https://openai.com/index/hello-gpt-4o/}}; open-weight models by Mistral (Mistral family), namely \texttt{open-mixtral-8x22b} (L), \texttt{open-mixtral-8x7b} (M), \texttt{open-mistral-7b} (S) \citep{jiang2024mixtralexperts}\footnote{\url{https://mistral.ai/news/mixtral-8x22b/}}; and open-weight models by Meta (Llama family), namely \texttt{llama-3.1-70b} (L), \texttt{llama-3.1-8b} (S) \citep{dubey2024llama3herdmodels}\footnote{\url{https://ai.meta.com/blog/meta-llama-3-1/}} .

We report the average classification accuracy (over the three random seeds) and the standard deviation for each LLM on the test subset of LAR-ECHR. 
We also show results on two previous legal benchmarks (LegalBench, MMLU-Law) and two general benchmarks (MMLU-full, HellaSwag), as previously reported \citep{liang2023holisticevaluationlanguagemodels}. 

MMLU \citep{hendryckstest2021} is the most widely used benchmark for evaluating the knowledge and reasoning abilities of instruction following LLMs \citep{liang2023holisticevaluationlanguagemodels}. MMLU-Law is a subset of MMLU that contains three legal tasks (`International Law', `Jurisprudence', `Professional Law'). LegalBench is the largest (in terms of tasks) benchmark for the evaluation of legal reasoning \citep{magesh2024hallucinationfreeassessingreliabilityleading}. It includes 162 tasks that assess 6 different reasoning types. HellaSwag \citep{zellers-etal-2019-hellaswag} is a dataset created automatically that only contains the next statement prediction task, similar to LAR-ECHR. However, in HellaSwag the texts are collected from online articles and not chains of legal arguments, as in LAR-ECHR. In the three previous benchmarks that have multiple tasks (MMLU, MMLU-Law, LegalBench), we report macro-average over their tasks.


\subsection{Zero-shot prompting}
All experiments are performed in a zero-shot setting with a Chain-of-Thought (CoT) prompt \citep{wei2023chainofthoughtpromptingelicitsreasoning} we designed for LAR-ECHR (Table \ref{tab:prompt}). The prompt explains the provided input (facts, preceding arguments, continuation options) and the task, asking the LLM to generate an output in three steps. In the first step (`Analysis'), the LLM reflects on the plausibility of each option (candidate next argument). In the second step (`Explanation'), the LLM explains its choice. In the last step (`Answer'), the LLM outputs only the letter (A, B, C, D) of its choice (to facilitate answer collection). 

\subsection{Experimental results}
\label{sec:results}

Table \ref{tab:main} presents the accuracy of the models on the five datasets (LAR-ECHR, LegalBench, MMLU-Law, MMLU-Full, HellaSwag). The large version of \texttt{GPT-4o} (L) achieves the best performance on all datasets, with accuracy 75.8\% on LAR-ECHR, showing that there is room for model improvement on our dataset. The rankings (in square brackets) of all models on LAR-ECHR  are identical to those of LegalBench, even though the two datasets are based on different legal systems (US, EU). Within each family of models, the largest model is the best on all datasets, as expected. 
\begin{table}[t!]
  \centering
  \small
  \renewcommand{\arraystretch}{1.1} 
  \begin{tabular}{|l|}
    \hline
    \textbf{Prompt:}\\
    You will be provided with the introductory Facts\\
    in a European Court of Human Rights (ECHR)\\
    case, an excerpt of arguments from that case and\\
    several possible continuations of these arguments.\\
    Your task is to determine which continuation \\
    accurately extends the original argument.\\
    Output format:\\
    - Analysis: Provide an in-depth evaluation\\ of how each option relates to the initial argument.\\
    - Explanation: Articulate why your chosen\\ option is the correct continuation.\\
    - Answer: State the letter of your selected answer.\\
    \textbf{Facts: \quad . . .} \\
    \textbf{Preceding arguments: \quad . . .}\\
    \textbf{Continuation options: \quad . . .}\\
    \textbf{LLM response:}\\
    \texttt{\textbf{Analysis:}}\\
    \texttt{(A): Reinforces B's mental health issues}\\
    \texttt{and need for treatment.}\\
    \texttt{(B): Discusses mental hospital confinement,}\\ 
    \texttt{unrelated to specific case facts.}\\
    \texttt{(C): Introduces irrelevant information}\\ 
    \texttt{about police actions and examinations.}\\
    \texttt{(D): Diverges to complaints about forced}\\ 
    \texttt{evaluations, not B's violence.}\\
    \texttt{\textbf{Explanation:}}\\
    \texttt{Option (A) criticizes Croatian authorities}\\
    \texttt{for not acting on B's known psychiatric}\\  
    \texttt{issues and violent tendencies, aligning }\\
    \texttt{with the Court's view of ongoing abuse.}\\
    \texttt{\textbf{Answer:} (A)}\\
    \hline
  \end{tabular}
  \caption{Our CoT prompt for LAR-ECHR, a sample input (facts, preceding arguments, continuation options), and the response (analysis, explanation, answer) from \texttt{GPT-4o}. The `...' are as in Table \ref{tab:example}. The full version of this table can be found in Appendix~\ref{sec:appendix-prompt}.}
  \label{tab:prompt}
\end{table}

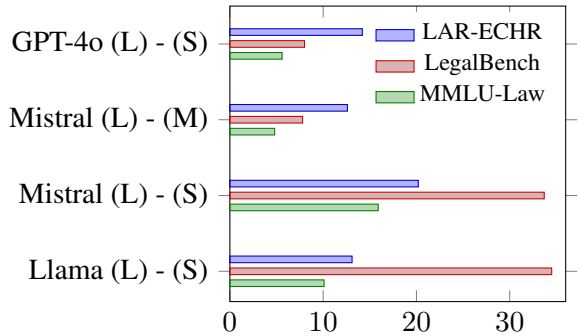
\begin{figure}[t]
\centering
\begin{tikzpicture}

\begin{axis} [xbar = 0.07cm,
    bar width = 2.5pt,
    xmin = 0,
    xmax = 36,
    ymin = -0.4,
    ymax = 3.5,
    ytick = {0, 1, 2, 3},
    yticklabels = {Llama (L) - (S), Mistral (L) - (S), Mistral (L) - (M), GPT-4o (L) - (S)},
    legend style={draw=none},
    width = 6cm,
    height = 5.5cm,
    reverse legend=true,
    legend image code/.code={ \draw[#1] (0cm,-0.1cm) rectangle (0.5cm,0.05cm);
                }
]

\addplot[green!50!black,fill=green!50!black,fill opacity=0.3] coordinates {(10.1,0) (15.9,1) (4.8,2) (5.6,3)};

\addplot[red!80!black,fill=red!50!black,fill opacity=0.3] coordinates {(34.5,0) (33.7,1) (7.8,2) (8.0,3)};
\addplot[blue!80!black,fill=blue,fill opacity=0.3] coordinates {(13.1,0) (20.2,1) (12.6,2) (14.2,3)};

\legend{{\small MMLU-Law},{\small LegalBench},{\small LAR-ECHR}}

\end{axis}

\end{tikzpicture}
\caption{Performance gap within the same LLM family.}
\label{fig:gap}
\end{figure}

\texttt{Mistral} (L) is the second best model on LAR-ECHR and LegalBench, with a larger performance gap from the best model (6 pp), compared to the corresponding gap on LegalBench (2.5 pp), showing that LAR-ECHR distinguishes better than LegalBench the legal reasoning abilities of the top two models. When comparing models of the same family, LAR-ECHR also distinguishes better between \texttt{GPT-4o} (L) and \texttt{GPT-4o} (S), as shown in Fig.~\ref{fig:gap}, and the same applies between \texttt{Mistral} (L) and \texttt{Mistral} (M). By contrast, LegalBench distinguishes substantially better between \texttt{Mistral} (L) and \texttt{Mistral} (S) (Fig.~\ref{fig:gap}), as well as between \texttt{Llama} (L) and \texttt{Llama} (S); this is due to the much lower scores the smaller \texttt{Mistral} (S) and \texttt{Llama} (S) obtain on LegalBench compared to LAR-ECHR.

Table~\ref{tab:summary} presents the performance of \texttt{GPT} and \texttt{Llama} models on LAR-ECHR when provided with the original, complete facts of the proceedings, for models with large enough context length. As expected, all models exhibit greater performance when using the complete facts. Notably, \texttt{Llama} models benefit more than \texttt{GPT} models. Differences between models within the same family are relatively small. These findings suggest that while summaries offer an effective workaround for models with limited context lengths, they can introduce bias, potentially favoring certain models. 

\section{Related work} \label{sec:related}
The LAR task was inspired by the continuation task introduced by SWAG \citep{zellers-etal-2018-swag} and later improved by HellaSwag \citep{zellers-etal-2019-hellaswag}.
It is a multiple-choice task where the model has to select the most likely continuation of an event description, such as ``A woman sits at a piano'' is followed by ``She sets her fingers on the keys''. The corpus is collected from various online sources such as wikiHow\footnote{\url{https://www.wikihow.com/}}. Similarly to LAR,  HellaSwag is constructed automatically, via a technique called Adversarial Filtering (AF) which selects the most persuasive LLM-generated continuations as incorrect options. It is shown empirically that accurately predicting the correct continuation of an event in HellaSwag requires skills that are closely related to commonsense reasoning. 
The primary differences with our work, aside from our focus on the legal domain, are: (a) we employ official content from court proceedings instead of events from online articles of varying credibility, (b) we use (based on the respective annotations) the most appropriate chain of arguments, and (c) we utilize human-generated challenging distractors.

\begin{table}[t]
  \centering
  \begin{tabular}{l|c}
    Models & \, LAR-ECHR \\
    &(complete facts)\\
    \hline
    GPT-4o \, & \textbf{77.9} (+2.1)\\
    GPT-4o-mini & 64.3 (+2.7) \\
    \hline
    Llama-70B & 73.3 (+6.1)  \\
    Llama-8B  & 58.0 (+3.9)   \\
    \hline
  \end{tabular}
  \caption{\label{tab:summary}
    Results on LAR-ECHR with complete facts. The difference in performance from the summarized version is shown in parentheses.
  }
\end{table}

Our dataset builds on top of two previous works: LAM:ECHR \citep{habernal_mining_2024} and ECtHR~B \citep{chalkidis-etal-2021-paragraph}.
LAM:ECHR annotated, with the help of legal experts, the arguments of 373 ECHR decisions with \textit{actor} and \textit{argument type} labels, and trained and evaluated their RoBERTa-based models on both tasks.
In ECtHR B the goal is to predict the articles of ECHR that were allegedly violated, given the facts of the case. 
To create LAR-ECHR we aligned the common instances of these datasets to combine the annotated arguments of LAM:ECHR with the facts of the cases from ECtHR~B.

One of the most widely known benchmarks in legal NLP is LexGLUE \citep{chalkidis-etal-2022-lexglue}. It was one of the first large-scale collection of datasets dedicated to the legal domain.
Its creation was inspired by the success of GLUE, a multi-task benchmark dataset \citep{wang-etal-2018-glue}, and the subsequent and more challening SuperGLUE \citep{NEURIPS2019_4496bf24}. 
LexGLUE includes a variety of (English-only) classification tasks from both US and EU legal systems, however it does not contain any reasoning-specific tasks.
LEXTREME \citep{niklaus-etal-2023-lextreme} followed with a collection of 11 datasets, featuring tasks similar to those in LexGLUE, to establish a multilingual legal NLP benchmark.
LegalLens \citep{bernsohn-etal-2024-legallens} introduced two classification tasks: detecting legal violations and identifying potentially affected individuals. The tasks were created using LLMs and then validated by human experts.
The aforementioned benchmarks focus on specific classification tasks. They do not directly measure in-context learning capabilities or the understanding of legal reasoning explanations. 

In the broader NLP landscape, several datasets have recently emerged for evaluating the few-shot learning capabilities and advanced reasoning skills of LLMs, replacing GLUE and SuperGLUE as the most widely used benchmarks.
These new benchmarks are more aligned with the skills required by chatbot assistants designed to solve a wide range of tasks by following instructions, primarily through generating text rather than predefined labels.
The most prominent of these is the Massive Multitask Language Understanding (MMLU) benchmark \citep{hendryckstest2021}, which is preferred for evaluating the knowledge and general capabilities of LLMs \citep{liang2023holisticevaluationlanguagemodels}.
It is a multiple-choice dataset that covers 57 tasks across diverse academic subjects, three of them being law-specific. 
ARC \citep{clark2018thinksolvedquestionanswering} is another multiple-choice question-answering dataset that includes science questions requiring various types of reasoning.
Big Bench \citep{srivastava2023imitationgamequantifyingextrapolating} is a challenging dataset of 204 tasks that focuses on various topics among them arithmetic, logical, common-sense and algorithmic reasoning as well as language understanding and world knowledge.

Inspired by the success of these benchmarks, several benchmarks for the legal domain, with the same orientation, were also made available. For example, the largest legal reasoning benchmark (in terms of number of tasks) is LegalBench \citep{guha2023legalbench}, comprising 162 tasks that cover six different types of legal reasoning and focus on the US legal system. 
\citet{bongard-etal-2022-legal} created a challenging legal reasoning dataset by adapting questions from a textbook on US civil procedure which however is cast as a binary classification task and does not focus on continuations such as LAR-ECHR.
A few datasets that are focused on other legal systems than the US were also made available. For example, LawBench \citep{fei2023lawbench} consists of 20 tasks on Chinese law that evaluate legal knowledge understanding of LLMs. IL-TUR \citep{joshi-etal-2024-il} covers a wide range of multilingual legal text understanding and reasoning tasks for English and 9 Indian languages. Our dataset, LAR-ECHR, differs from the datasets in these benchmarks in that (a) it uses the legal reasoning chain of the arguments of the judges, (b) it refers to EU law and (c) instances are collected semi-automatically from court proceedings using annotations, not handcrafted by legal experts.

\section{Conclusion} 
\label{sec:conclusion}
In this study, we introduced LAR, a legal reasoning NLP task that requires selecting the correct next argument made by judges in a case.
We constructed a dataset for this task, called LAR-ECHR, using cases from ECHR.
We evaluated seven general-purpose LLMs from three families on this dataset.
The best model obtained 75.8\% accuracy, indicating significant potential for further model improvement.
Model rankings were identical with those of LegalBench, even though the datasets are based on different legal systems.
Despite that weak models obtained a substantially lower score in LegalBench, LAR-ECHR distinguished the top models more clearly.
The process followed to construct LAR-ECHR can be replicated with cases from any court proceedings, even from different legal systems.

\section{Future work}
\label{sec:future}

The semi-automatic creation of a LAR dataset requires a few design decisions, two of which we believe are most worth investigating further: (a) the impact of not selecting only the first arguments of the judges (per issue) as target arguments (which in our experience are the most difficult to predict) and (b) the impact of the similarity threshold $\tau$ in selecting candidate distractors.

Additionally, we plan to extend the dataset in various directions: (a) collect and align the missing ECHR cases that are annotated from LAM:ECHR, but they do not exist in ECtHR B, (b) include the rest of the articles of ECHR, apart from articles 7 and 8, to cover other domains of legal expertise, (c) annotate more cases to increase the dataset size.
These extensions could lead to the inclusion of a training set for fine-tuning LLMs. These LLMs would be either open-source LLMs or smaller BERT-based models that have shown promise in legal reasoning tasks, such as \citep{chalkidis-etal-2020-legal}. 
Even though these legal-specific models do not exhibit few-shot learning capabilities, they would be ideal baseline models.

Independently of this extension, we plan to evaluate more general-purpose, but also legal-specific LLMs, and update the leaderboard of the dataset.
It would be insightful to measure the impact of pretraining on the same or other legal systems.
To our knowledge, there is currently only one publicly released family of legal LLMs that can follow instructions, \texttt{Saul-7B} \citep{colombo2024saullm7bpioneeringlargelanguage}, \texttt{Saul-54B} and \texttt{Saul-141B} \citep{colombo2024saullm54bsaullm141bscaling}.

Finally, the process followed to construct LAR-ECHR could be replicated with cases from other court proceedings to create new LAR datasets that are focused on other legal systems and/or languages.

\section*{Limitations}
\label{sec:limitations}

One limitation of our work has to do with the process followed to create the dataset.  While the data were originally created by humans, the next statement prediction task is artificial. We employed semi-automatic techniques, based on legal expert annotations and embedding similarity of the arguments, to compile a challenging dataset. We also summarized the facts to fit in the context length of all the models. This process might have introduced biases and/or mistakes, as we have already discussed for the summaries of the facts (\textsection~\ref{sec:results}). The impact of these biases could only be measured by careful examination from legal experts and extensive comparisons with different variations (e.g. summaries from other models).

Furthermore, it should be noted that, as in many other legal NLP datasets, we are using the `facts' of ECHR court decisions as if they are the factual information available prior to the final decision. However, due to the details of the legal process and the way that court proceedings are written, this is unrealistic \citep{medvedeva-mcbride-2023-legal}. The judges actually publish only the information that is supporting their final decision as the `facts' of the case; not the original record that they had to consider in that process. To make the task realistic for a real-world application we should include the actual information that the parties had access to before the final judgement took place, but access to this information is very hard to get for most cases.

\section*{Ethics Statement} 
\label{sec:ethics}

The primary objective of this research is to advance legal NLP and more specifically the use of LLMs as tools that assist--without replacing--legal professionals.
A diverse set of communities can be benefited from our research:
(a) the NLP community can challenge existing and future LLMs on an advanced legal reasoning dataset and even build new datasets for other courts, (b) legal practitioners can improve their understanding of the way these models make decisions and (c) the legal tech community can gain useful insights into LLM capabilities across different courts and legal systems, enabling them to design appropriate use cases and develop more accurate tools.

Most previous work in legal NLP, including both benchmarks and models, \citep{guha2023legalbench, niklaus-etal-2023-lextreme, chalkidis-etal-2020-legal}, advocate that they do not aim to replace judges, but instead to assist them in reaching more informed decisions. However, most of them are trying to predict the outcome of legal decisions, without providing or evaluating legal reasoning explanations. In contrast, our work evaluates the ability of LLMs to identify the correct next statement in a judge's chain of legal arguments, which is closely linked to their capacity to produce valid legal reasoning.

When introducing a new legal NLP task, it is vital to consider the intended use cases for potential models designed for it \citep{medvedeva-mcbride-2023-legal, tsarapatsanis-aletras-2021-ethical}. In our case, we advocate that such models be used solely as supporting tools to review the reasoning of legal professionals, rather than to produce their own legal reasoning (let alone predict the outcome of a case).

For example, we propose developing a legal verification tool, i.e. a tool that can verify the validity of the legal reasoning of an argument chain. This tool could be used by judges to validate the `Law' section of the proceedings (after the final decision is taken) before publishing them. If a potential reasoning weakness is located by the model, then it could provide its own CoT explanation to pinpoint the root cause of the problem. The judges would then evaluate if they agree with the model or not, and if their reasoning requires revision. In this example ethical risks are almost completely mitigated, because the decision is already taken. This tool would help the judges prepare the proceedings faster and it could decrease oversights.







We recognize the ethical importance of data privacy and confidentiality.
All data is obtained from publicly accessible online sources, without infringing any proprietary rights, and in accordance with the licenses under which they were released.
The data from LAM:ECHR were released under the \texttt{`Apache 2.0'} license along with the respective software.\footnote{\url{https://github.com/trusthlt/mining-legal-arguments/blob/main/LICENSE}}
The data from ECtHR~B were released under the `Creative Commons Attribution-NonCommercial-ShareAlike  4.0' (\texttt{`CC BY-NC-SA 4.0'}) license.\footnote{\url{https://huggingface.co/datasets/AUEB-NLP/ecthr_cases}}
In accordance to \texttt{`CC BY-NC-SA 4.0'}, we released our dataset under the same license as well.\footnote{\url{https://creativecommons.org/licenses/by-nc-sa/4.0/}}

\section*{Acknowledgments}
This work has been partially supported by project MIS 5154714 of the National Recovery and Resilience Plan Greece 2.0 funded by the European Union under the NextGenerationEU Program. In addition, AWS resources were provided by the National Infrastructures for Research and Technology GRNET and funded by the EU Recovery and Resiliency Facility.

\bibliography{anthology,custom}

\begin{thebibliography}{31}
\providecommand{\natexlab}[1]{#1}

\bibitem[{Aletras et~al.(2016)Aletras, Tsarapatsanis, Preotiuc-Pietro, and Lampos}]{Aletras2016PredictingJD}
Nikolaos Aletras, Dimitrios Tsarapatsanis, Daniel Preotiuc-Pietro, and Vasileios Lampos. 2016.
\newblock Predicting judicial decisions of the european court of human rights: a natural language processing perspective.
\newblock \emph{PeerJ Comput. Sci.}, 2:e93.

\bibitem[{Bernsohn et~al.(2024)Bernsohn, Semo, Vazana, Hayat, Hagag, Niklaus, Saha, and Truskovskyi}]{bernsohn-etal-2024-legallens}
Dor Bernsohn, Gil Semo, Yaron Vazana, Gila Hayat, Ben Hagag, Joel Niklaus, Rohit Saha, and Kyryl Truskovskyi. 2024.
\newblock \href {https://aclanthology.org/2024.eacl-long.130} {{L}egal{L}ens: Leveraging {LLM}s for legal violation identification in unstructured text}.
\newblock In \emph{Proceedings of the 18th Conference of the European Chapter of the Association for Computational Linguistics (Volume 1: Long Papers)}, pages 2129--2145, St. Julian{'}s, Malta. Association for Computational Linguistics.

\bibitem[{Bongard et~al.(2022)Bongard, Held, and Habernal}]{bongard-etal-2022-legal}
Leonard Bongard, Lena Held, and Ivan Habernal. 2022.
\newblock \href {https://doi.org/10.18653/v1/2022.nllp-1.17} {The legal argument reasoning task in civil procedure}.
\newblock In \emph{Proceedings of the Natural Legal Language Processing Workshop 2022}, pages 194--207, Abu Dhabi, United Arab Emirates (Hybrid). Association for Computational Linguistics.

\bibitem[{Chalkidis et~al.(2020)Chalkidis, Fergadiotis, Malakasiotis, Aletras, and Androutsopoulos}]{chalkidis-etal-2020-legal}
Ilias Chalkidis, Manos Fergadiotis, Prodromos Malakasiotis, Nikolaos Aletras, and Ion Androutsopoulos. 2020.
\newblock \href {https://doi.org/10.18653/v1/2020.findings-emnlp.261} {{LEGAL}-{BERT}: The muppets straight out of law school}.
\newblock In \emph{Findings of the Association for Computational Linguistics: EMNLP 2020}, pages 2898--2904, Online. Association for Computational Linguistics.

\bibitem[{Chalkidis et~al.(2021)Chalkidis, Fergadiotis, Tsarapatsanis, Aletras, Androutsopoulos, and Malakasiotis}]{chalkidis-etal-2021-paragraph}
Ilias Chalkidis, Manos Fergadiotis, Dimitrios Tsarapatsanis, Nikolaos Aletras, Ion Androutsopoulos, and Prodromos Malakasiotis. 2021.
\newblock \href {https://doi.org/10.18653/v1/2021.naacl-main.22} {Paragraph-level rationale extraction through regularization: A case study on {E}uropean court of human rights cases}.
\newblock In \emph{Proceedings of the 2021 Conference of the North American Chapter of the Association for Computational Linguistics: Human Language Technologies}, pages 226--241, Online. Association for Computational Linguistics.

\bibitem[{Chalkidis et~al.(2022)Chalkidis, Jana, Hartung, Bommarito, Androutsopoulos, Katz, and Aletras}]{chalkidis-etal-2022-lexglue}
Ilias Chalkidis, Abhik Jana, Dirk Hartung, Michael Bommarito, Ion Androutsopoulos, Daniel Katz, and Nikolaos Aletras. 2022.
\newblock \href {https://doi.org/10.18653/v1/2022.acl-long.297} {{L}ex{GLUE}: A benchmark dataset for legal language understanding in {E}nglish}.
\newblock In \emph{Proceedings of the 60th Annual Meeting of the Association for Computational Linguistics (Volume 1: Long Papers)}, pages 4310--4330, Dublin, Ireland. Association for Computational Linguistics.

\bibitem[{Clark et~al.(2018)Clark, Cowhey, Etzioni, Khot, Sabharwal, Schoenick, and Tafjord}]{clark2018thinksolvedquestionanswering}
Peter Clark, Isaac Cowhey, Oren Etzioni, Tushar Khot, Ashish Sabharwal, Carissa Schoenick, and Oyvind Tafjord. 2018.
\newblock \href {https://arxiv.org/abs/1803.05457} {Think you have solved question answering? try arc, the ai2 reasoning challenge}.
\newblock \emph{Preprint}, arXiv:1803.05457.

\bibitem[{Colombo et~al.(2024{\natexlab{a}})Colombo, Pires, Boudiaf, Melo, Culver, Morgado, Malaboeuf, Hautreux, Charpentier, and Desa}]{colombo2024saullm54bsaullm141bscaling}
Pierre Colombo, Telmo Pires, Malik Boudiaf, Rui Melo, Dominic Culver, Sofia Morgado, Etienne Malaboeuf, Gabriel Hautreux, Johanne Charpentier, and Michael Desa. 2024{\natexlab{a}}.
\newblock \href {https://arxiv.org/abs/2407.19584} {Saullm-54b {\&} saullm-141b: Scaling up domain adaptation for the legal domain}.
\newblock \emph{Preprint}, arXiv:2407.19584.

\bibitem[{Colombo et~al.(2024{\natexlab{b}})Colombo, Pires, Boudiaf, Culver, Melo, Corro, Martins, Esposito, Raposo, Morgado, and Desa}]{colombo2024saullm7bpioneeringlargelanguage}
Pierre Colombo, Telmo~Pessoa Pires, Malik Boudiaf, Dominic Culver, Rui Melo, Caio Corro, Andre F.~T. Martins, Fabrizio Esposito, Vera~Lúcia Raposo, Sofia Morgado, and Michael Desa. 2024{\natexlab{b}}.
\newblock \href {https://arxiv.org/abs/2403.03883} {Saullm-7b: A pioneering large language model for law}.
\newblock \emph{Preprint}, arXiv:2403.03883.

\bibitem[{Dalvi et~al.(2021)Dalvi, Jansen, Tafjord, Xie, Smith, Pipatanangkura, and Clark}]{dalvi-etal-2021-explaining}
Bhavana Dalvi, Peter Jansen, Oyvind Tafjord, Zhengnan Xie, Hannah Smith, Leighanna Pipatanangkura, and Peter Clark. 2021.
\newblock \href {https://doi.org/10.18653/v1/2021.emnlp-main.585} {Explaining answers with entailment trees}.
\newblock In \emph{Proceedings of the 2021 Conference on Empirical Methods in Natural Language Processing}, pages 7358--7370, Online and Punta Cana, Dominican Republic. Association for Computational Linguistics.

\bibitem[{Dubey et~al.(2024)Dubey, Jauhri, Pandey, Kadian, Al-Dahle, Letman, Mathur, Schelten, Yang, Fan, Goyal, Hartshorn, Yang, Mitra, Sravankumar, Korenev, Hinsvark, Rao, Zhang, Rodriguez, Gregerson et~al.}]{dubey2024llama3herdmodels}
Abhimanyu Dubey, Abhinav Jauhri, Abhinav Pandey, Abhishek Kadian, Ahmad Al-Dahle, Aiesha Letman, Akhil Mathur, Alan Schelten, Amy Yang, Angela Fan, Anirudh Goyal, Anthony Hartshorn, Aobo Yang, Archi Mitra, Archie Sravankumar, Artem Korenev, Arthur Hinsvark, Arun Rao, Aston Zhang, Aurelien Rodriguez, Austen Gregerson, et~al. 2024.
\newblock \href {https://arxiv.org/abs/2407.21783} {The llama 3 herd of models}.
\newblock \emph{Preprint}, arXiv:2407.21783.

\bibitem[{Fei et~al.(2023)Fei, Shen, Zhu, Zhou, Han, Zhang, Chen, Shen, and Ge}]{fei2023lawbench}
Zhiwei Fei, Xiaoyu Shen, Dawei Zhu, Fengzhe Zhou, Zhuo Han, Songyang Zhang, Kai Chen, Zongwen Shen, and Jidong Ge. 2023.
\newblock Lawbench: Benchmarking legal knowledge of large language models.
\newblock \emph{arXiv preprint arXiv:2309.16289}.

\bibitem[{Guha et~al.(2023)Guha, Nyarko, Ho, Ré, Chilton, Narayana, Chohlas-Wood, Peters, Waldon, Rockmore, Zambrano, Talisman, Hoque, Surani, Fagan, Sarfaty, Dickinson, Porat, Hegland et~al.}]{guha2023legalbench}
Neel Guha, Julian Nyarko, Daniel~E. Ho, Christopher Ré, Adam Chilton, Aditya Narayana, Alex Chohlas-Wood, Austin Peters, Brandon Waldon, Daniel~N. Rockmore, Diego Zambrano, Dmitry Talisman, Enam Hoque, Faiz Surani, Frank Fagan, Galit Sarfaty, Gregory~M. Dickinson, Haggai Porat, Jason Hegland, et~al. 2023.
\newblock \href {https://arxiv.org/abs/2308.11462} {Legalbench: A collaboratively built benchmark for measuring legal reasoning in large language models}.
\newblock \emph{Preprint}, arXiv:2308.11462.

\bibitem[{Habernal et~al.(2024)Habernal, Faber, Recchia, Bretthauer, Gurevych, Spiecker~genannt D{\"o}hmann, and Burchard}]{habernal_mining_2024}
Ivan Habernal, Daniel Faber, Nicola Recchia, Sebastian Bretthauer, Iryna Gurevych, Indra Spiecker~genannt D{\"o}hmann, and Christoph Burchard. 2024.
\newblock \href {https://doi.org/10.1007/s10506-023-09361-y} {Mining legal arguments in court decisions}.
\newblock \emph{Artificial Intelligence and Law}, 32(3):1--38.

\bibitem[{Hendrycks et~al.(2021)Hendrycks, Burns, Basart, Zou, Mazeika, Song, and Steinhardt}]{hendryckstest2021}
Dan Hendrycks, Collin Burns, Steven Basart, Andy Zou, Mantas Mazeika, Dawn Song, and Jacob Steinhardt. 2021.
\newblock Measuring massive multitask language understanding.
\newblock \emph{Proceedings of the International Conference on Learning Representations (ICLR)}.

\bibitem[{Jiang et~al.(2024)Jiang, Sablayrolles, Roux, Mensch, Savary, Bamford, Chaplot, de~las Casas, Hanna, Bressand, Lengyel, Bour, Lample, Lavaud, Saulnier, Lachaux, Stock et~al.}]{jiang2024mixtralexperts}
Albert~Q. Jiang, Alexandre Sablayrolles, Antoine Roux, Arthur Mensch, Blanche Savary, Chris Bamford, Devendra~Singh Chaplot, Diego de~las Casas, Emma~Bou Hanna, Florian Bressand, Gianna Lengyel, Guillaume Bour, Guillaume Lample, Lélio~Renard Lavaud, Lucile Saulnier, Marie-Anne Lachaux, Pierre Stock, et~al. 2024.
\newblock \href {https://arxiv.org/abs/2401.04088} {Mixtral of experts}.
\newblock \emph{Preprint}, arXiv:2401.04088.

\bibitem[{Joshi et~al.(2024)Joshi, Paul, Sharma, Goyal, Ghosh, and Modi}]{joshi-etal-2024-il}
Abhinav Joshi, Shounak Paul, Akshat Sharma, Pawan Goyal, Saptarshi Ghosh, and Ashutosh Modi. 2024.
\newblock \href {https://aclanthology.org/2024.acl-long.618} {{IL}-{TUR}: Benchmark for {I}ndian legal text understanding and reasoning}.
\newblock In \emph{Proceedings of the 62nd Annual Meeting of the Association for Computational Linguistics (Volume 1: Long Papers)}, pages 11460--11499, Bangkok, Thailand. Association for Computational Linguistics.

\bibitem[{Liang et~al.(2023)Liang, Bommasani, Lee, Tsipras, Soylu, Yasunaga, Zhang, Narayanan, Wu, Kumar, Newman, Yuan, Yan, Zhang, Cosgrove, Manning, Ré et~al.}]{liang2023holisticevaluationlanguagemodels}
Percy Liang, Rishi Bommasani, Tony Lee, Dimitris Tsipras, Dilara Soylu, Michihiro Yasunaga, Yian Zhang, Deepak Narayanan, Yuhuai Wu, Ananya Kumar, Benjamin Newman, Binhang Yuan, Bobby Yan, Ce~Zhang, Christian Cosgrove, Christopher~D. Manning, Christopher Ré, et~al. 2023.
\newblock \href {https://arxiv.org/abs/2211.09110} {Holistic evaluation of language models}.
\newblock \emph{Preprint}, arXiv:2211.09110.

\bibitem[{Magesh et~al.(2024)Magesh, Surani, Dahl, Suzgun, Manning, and Ho}]{magesh2024hallucinationfreeassessingreliabilityleading}
Varun Magesh, Faiz Surani, Matthew Dahl, Mirac Suzgun, Christopher~D. Manning, and Daniel~E. Ho. 2024.
\newblock \href {https://arxiv.org/abs/2405.20362} {Hallucination-free? assessing the reliability of leading ai legal research tools}.
\newblock \emph{Preprint}, arXiv:2405.20362.

\bibitem[{Medvedeva and Mcbride(2023)}]{medvedeva-mcbride-2023-legal}
Masha Medvedeva and Pauline Mcbride. 2023.
\newblock \href {https://doi.org/10.18653/v1/2023.nllp-1.9} {Legal judgment prediction: If you are going to do it, do it right}.
\newblock In \emph{Proceedings of the Natural Legal Language Processing Workshop 2023}, pages 73--84, Singapore. Association for Computational Linguistics.

\bibitem[{Neelakantan et~al.(2022)Neelakantan, Xu, Puri, Radford, Han, Tworek, Yuan, Tezak, Kim, Hallacy, Heidecke, Shyam, Power, Nekoul, Sastry, Krueger, Schnurr, Such, Hsu, and othes}]{neelakantan2022textcodeembeddingscontrastive}
Arvind Neelakantan, Tao Xu, Raul Puri, Alec Radford, Jesse~Michael Han, Jerry Tworek, Qiming Yuan, Nikolas Tezak, Jong~Wook Kim, Chris Hallacy, Johannes Heidecke, Pranav Shyam, Boris Power, Tyna~Eloundou Nekoul, Girish Sastry, Gretchen Krueger, David Schnurr, Felipe~Petroski Such, Kenny Hsu, and othes. 2022.
\newblock \href {https://arxiv.org/abs/2201.10005} {Text and code embeddings by contrastive pre-training}.
\newblock \emph{Preprint}, arXiv:2201.10005.

\bibitem[{Niklaus et~al.(2023)Niklaus, Matoshi, Rani, Galassi, St{\"u}rmer, and Chalkidis}]{niklaus-etal-2023-lextreme}
Joel Niklaus, Veton Matoshi, Pooja Rani, Andrea Galassi, Matthias St{\"u}rmer, and Ilias Chalkidis. 2023.
\newblock \href {https://doi.org/10.18653/v1/2023.findings-emnlp.200} {{LEXTREME}: A multi-lingual and multi-task benchmark for the legal domain}.
\newblock In \emph{Findings of the Association for Computational Linguistics: EMNLP 2023}, pages 3016--3054, Singapore. Association for Computational Linguistics.

\bibitem[{OpenAI et~al.(2024)OpenAI, Achiam, Adler, Agarwal, Ahmad, Akkaya, Aleman, Almeida, Altenschmidt, Altman, Anadkat, Avila, Babuschkin, Balaji, Balcom, Baltescu, Bao, Bavarian, Belgum et~al.}]{openai2024gpt4technicalreport}
OpenAI, Josh Achiam, Steven Adler, Sandhini Agarwal, Lama Ahmad, Ilge Akkaya, Florencia~Leoni Aleman, Diogo Almeida, Janko Altenschmidt, Sam Altman, Shyamal Anadkat, Red Avila, Igor Babuschkin, Suchir Balaji, Valerie Balcom, Paul Baltescu, Haiming Bao, Mohammad Bavarian, Jeff Belgum, et~al. 2024.
\newblock \href {https://arxiv.org/abs/2303.08774} {Gpt-4 technical report}.
\newblock \emph{Preprint}, arXiv:2303.08774.

\bibitem[{Srivastava et~al.(2023)Srivastava, Rastogi, Rao, Shoeb, Abid, Fisch, Brown, Santoro, Gupta, Garriga-Alonso, Kluska, Lewkowycz, Agarwal, Power, Ray, Warstadt, Kocurek, Safaya,  et~al.}]{srivastava2023imitationgamequantifyingextrapolating}
Aarohi Srivastava, Abhinav Rastogi, Abhishek Rao, Abu Awal~Md Shoeb, Abubakar Abid, Adam Fisch, Adam~R. Brown, Adam Santoro, Aditya Gupta, Adrià Garriga-Alonso, Agnieszka Kluska, Aitor Lewkowycz, Akshat Agarwal, Alethea Power, Alex Ray, Alex Warstadt, Alexander~W. Kocurek, Ali Safaya, , et~al. 2023.
\newblock \href {https://arxiv.org/abs/2206.04615} {Beyond the imitation game: Quantifying and extrapolating the capabilities of language models}.
\newblock \emph{Preprint}, arXiv:2206.04615.

\bibitem[{Tsarapatsanis and Aletras(2021)}]{tsarapatsanis-aletras-2021-ethical}
Dimitrios Tsarapatsanis and Nikolaos Aletras. 2021.
\newblock \href {https://doi.org/10.18653/v1/2021.findings-acl.314} {On the ethical limits of natural language processing on legal text}.
\newblock In \emph{Findings of the Association for Computational Linguistics: ACL-IJCNLP 2021}, pages 3590--3599, Online. Association for Computational Linguistics.

\bibitem[{T.y.s.s. et~al.(2024)T.y.s.s., Haddad, and Grabmair}]{t-y-s-s-etal-2024-ecthr}
Santosh T.y.s.s., Rashid Haddad, and Matthias Grabmair. 2024.
\newblock \href {https://aclanthology.org/2024.lrec-main.486} {{EC}t{HR}-{PCR}: A dataset for precedent understanding and prior case retrieval in the {E}uropean court of human rights}.
\newblock In \emph{Proceedings of the 2024 Joint International Conference on Computational Linguistics, Language Resources and Evaluation (LREC-COLING 2024)}, pages 5473--5483, Torino, Italia. ELRA and ICCL.

\bibitem[{Wang et~al.(2019)Wang, Pruksachatkun, Nangia, Singh, Michael, Hill, Levy, and Bowman}]{NEURIPS2019_4496bf24}
Alex Wang, Yada Pruksachatkun, Nikita Nangia, Amanpreet Singh, Julian Michael, Felix Hill, Omer Levy, and Samuel Bowman. 2019.
\newblock \href {https://proceedings.neurips.cc/paper_files/paper/2019/file/4496bf24afe7fab6f046bf4923da8de6-Paper.pdf} {Superglue: A stickier benchmark for general-purpose language understanding systems}.
\newblock In \emph{Advances in Neural Information Processing Systems}, volume~32. Curran Associates, Inc.

\bibitem[{Wang et~al.(2018)Wang, Singh, Michael, Hill, Levy, and Bowman}]{wang-etal-2018-glue}
Alex Wang, Amanpreet Singh, Julian Michael, Felix Hill, Omer Levy, and Samuel Bowman. 2018.
\newblock \href {https://doi.org/10.18653/v1/W18-5446} {{GLUE}: A multi-task benchmark and analysis platform for natural language understanding}.
\newblock In \emph{Proceedings of the 2018 {EMNLP} Workshop {B}lackbox{NLP}: Analyzing and Interpreting Neural Networks for {NLP}}, pages 353--355, Brussels, Belgium. Association for Computational Linguistics.

\bibitem[{Wei et~al.(2023)Wei, Wang, Schuurmans, Bosma, Ichter, Xia, Chi, Le, and Zhou}]{wei2023chainofthoughtpromptingelicitsreasoning}
Jason Wei, Xuezhi Wang, Dale Schuurmans, Maarten Bosma, Brian Ichter, Fei Xia, Ed~Chi, Quoc Le, and Denny Zhou. 2023.
\newblock \href {https://arxiv.org/abs/2201.11903} {Chain-of-thought prompting elicits reasoning in large language models}.
\newblock \emph{Preprint}, arXiv:2201.11903.

\bibitem[{Zellers et~al.(2018)Zellers, Bisk, Schwartz, and Choi}]{zellers-etal-2018-swag}
Rowan Zellers, Yonatan Bisk, Roy Schwartz, and Yejin Choi. 2018.
\newblock \href {https://doi.org/10.18653/v1/D18-1009} {{SWAG}: A large-scale adversarial dataset for grounded commonsense inference}.
\newblock In \emph{Proceedings of the 2018 Conference on Empirical Methods in Natural Language Processing}, pages 93--104, Brussels, Belgium. Association for Computational Linguistics.

\bibitem[{Zellers et~al.(2019)Zellers, Holtzman, Bisk, Farhadi, and Choi}]{zellers-etal-2019-hellaswag}
Rowan Zellers, Ari Holtzman, Yonatan Bisk, Ali Farhadi, and Yejin Choi. 2019.
\newblock \href {https://doi.org/10.18653/v1/P19-1472} {{H}ella{S}wag: Can a machine really finish your sentence?}
\newblock In \emph{Proceedings of the 57th Annual Meeting of the Association for Computational Linguistics}, pages 4791--4800, Florence, Italy. Association for Computational Linguistics.

\end{thebibliography}
\appendix


\section{Prompt}
\label{sec:appendix-prompt}


The following is the complete version of the prompt presented in Table~\ref{tab:prompt}, that we designed for \mbox{LAR-ECHR}.
\begin{lstlisting}
You will be provided with the introductory Facts in a European Court of Human Rights (ECHR) case, an excerpt of arguments of that case and several possible continuations of these arguments. Your task is to determine which continuation accurately extends the original argument from the case. To complete this task successfully:

1. Thoroughly analyze each provided option to identify its connection to the initial argument presented.
2. Choose the option that not only maintains the theme and context of the initial argument but also follows logically and seamlessly from it.
3. After selecting the most appropriate continuation, provide a detailed rationale for your choice.
4. Clearly state your answer by specifying the letter corresponding to the correct option.

Output format:
- Analysis: Provide an in-depth evaluation of how each option relates to the initial argument.
- Explanation: Articulate why your chosen option is the correct continuation.
- Answer: State the letter of your selected answer.
\end{lstlisting}


\section{Complete Example}
\label{sec:full-examples}
The following is the complete version of the instance presented in Table \ref{tab:example}. The facts of the case are presented first, then the examples as they were given to the LLMs for evaluation. After the input of the example, the complete CoT response of \texttt{GPT-4o} follows to help the reader understand the reasoning of the LLM. The explanations should not be considered an accurate representation of the inner thinking of the LLM, but still can help us understand its reasoning and can help non-experts understand the legal terms.\\

\noindent\textbf{Facts:}
\begin{lstlisting}
The applicant, born in 1979, married B in April 2001, giving birth to a daughter, C, shortly after. B, who suffered from psychiatric disorders due to his traumatic experiences during the Homeland War, subjected the applicant to verbal and physical abuse between 2003 and 2005. He was arrested in November 2005 and indicted for violent behavior but released in December 2005. The applicant moved to a women's shelter with C in January 2006 for safety. B continued his abusive behavior, leading to further legal proceedings, including charges of making death threats against the applicant and a police officer, for which he was found guilty and sentenced to imprisonment in 2006. A restraining order was issued, and subsequent appeals by B and the State Attorney's Office were dismissed in 2007, though the sentence was not enforced. Multiple other criminal and minor offenses cases were initiated against B over the years for domestic violence and threatening behavior. Meanwhile, the marriage of the applicant and B was dissolved in November 2006. The applicant faced difficulties in securing safe accommodation due to B's continued harassment, including hiring a private detective to locate her. The legal proceedings against B were ongoing, with several hearings adjourned due to B's absence, and no psychiatric treatment had been ensured despite recommendations. Overall, the applicant struggled with legal enforcement and protective measures against B's continued threat and harassment, affecting her and her daughter's safety and stability.    
\end{lstlisting}

\noindent\textbf{Arguments:}
\begin{lstlisting}
The Government argued that in Croatia the protection of victims of domestic violence was ensured through the mechanisms of criminal law, and in particular the Protection against Domestic Violence Act. In the present case the relevant authorities had reacted to the incidents of violence against the applicant by B, had instituted several sets of both criminal and minor offences proceedings and had applied such criminal sanctions and protective measures against B as they had considered proper and suitable in the circumstances. The Government submitted that the prison term imposed on B for not paying in full the fine imposed in the decision of the Z. Minor Offences Court of 2 October 2006 had not been enforced because Z. Prison had been full to capacity. Likewise, the measure of compulsory psycho-social treatment imposed on B in the same decision had not been implemented owing to the lack of licensed individuals or agencies able to execute such a protective measure (see paragraphs 31 and 34 above).
In addition, the Government had adopted two national strategies for protection against domestic violence (the first one covering the period between 2005 and 2007 and the second covering the period between 2008 and 2010) which included, inter alia, the education of all those involved in cases of domestic violence and cooperation with the non-governmental organisations working in that field as well as financial and other support for them. Thus, in 2008 only sixteen new shelters with a total of 329 places for the victims of violence had been established, of which six were State-funded.
The Court takes note of B's repeated violent behaviour towards the applicant. The facts in issue concern frequent episodes of violence in the period between November 2003 and June 2006, amounting to some two years and seven months. The violence was both verbal, including serious death threats, and physical, including hitting and kicking the applicant in the head, face and body, causing her injuries. In view of the fact that all the incidents of domestic violence in the present case concerned the same perpetrator and occurred in a continual manner, the Court will examine them as a continuous situation.
\end{lstlisting}
\textbf{Continuation options:}
\begin{lstlisting}

(A) The Court takes further note of the psychiatric reports concerning B which indicated that he suffered from several mental disorders, including a severe form of PTSD, emphasised his tendency towards violence and his reduced ability to control his impulses, and reiterated the recommendation for continuing compulsory psychiatric treatment (see paragraphs 6, 12 and 13 above).

(B) Insofar as the applicant's complaint under Article 5 - 1 of the Convention is concerned, his confinement in the mental hospital can be divided into two periods: between 9 and 20 December 2002, and after 20 December 2002, when the hospital became his guardian.

(C) The Court notes, and this is not in dispute, that in August and October 2009 the police collected information about the applicant's character from his neighbours and relatives, pursuant to the instruction of the President of the Krasnogvardiyskyy Court (see paragraphs 13-15 and 20 above). That information was apparently used by the psychiatric hospital's specialists in the context of the forensic psychiatric examination of the applicant.

(D) The applicant firstly complained that (i) the decisions to order her to undergo a psychiatric examination and to arrest her and (ii) her detention for a total period of eighty-three days were in breach of Article 8 of the Convention. However, the Court considers that these complaints do not raise any separate issue which is not covered by the finding of a violation of Article 5 - 1 of the Convention.
\end{lstlisting}

\noindent\textbf{GPT-4o Response:}
\\

\noindent\texttt{Analysis:}\\
\texttt{To find the correct continuation of the initial argument, we need to focus on both the factual context and legal analysis provided. }\\
\texttt{- Option (A) aligns closely with the initial arguments by addressing the psychiatric reports about B, reinforcing the situation’s consistency with his diagnosed mental disorders. It maintains the theme of continuous violence and the need for psychiatric treatment, linking back to the facts concerning B's mental health.
}\\
\texttt{- Option (B) discusses periods of confinement in a mental hospital, but this appears unrelated to the specific facts of the case involving B’s violent behavior toward the applicant and does not further the argument regarding domestic violence or its handling by Croatian authorities.
}\\
\texttt{- Option (C) mentions the police collecting information about the applicant’s character and a forensic psychiatric examination, which appears out of context, as there is no mention of such actions linked to either the applicant or B in the provided facts. 
}\\
\texttt{- Option (D) pertains to complaints about forced psychiatric evaluations and unrelated detention, diverging from the continuity of violence by B and responses by Croatian authorities, and thus, it does not logically follow the given argument.
}\\
\\
\noindent\texttt{Explanation:}\\
\texttt{Option (A) is the most logical continuation as it directly ties into and extends the argument concerning the failure of the Croatian authorities to implement appropriate protective measures and treatment for B despite his documented psychiatric issues and violent tendencies. This reflects the Court's approach of understanding the situation as a "continuous situation" of abuse, emphasizing the importance of ongoing psychiatric intervention, which was previously recommended and not enacted.}\\
\\
\texttt{Answer: A} \quad \faCheck

\end{document}